\documentclass[sigconf,scrartcl, 9pt]{acmart}

\AtBeginDocument{%
  \providecommand\BibTeX{{%
    \normalfont B\kern-0.5em{\scshape i\kern-0.25em b}\kern-0.8em\TeX}}}

\copyrightyear{2020}
\acmYear{2020}
\setcopyright{acmcopyright}\acmConference[ICCAD '20]{IEEE/ACM International Conference on Computer-Aided Design}{November 2--5, 2020}{Virtual Event, USA}
\acmBooktitle{IEEE/ACM International Conference on Computer-Aided Design (ICCAD '20), November 2--5, 2020, Virtual Event, USA}
\acmPrice{15.00}
\acmDOI{10.1145/3400302.3415675}
\acmISBN{978-1-4503-8026-3/20/11}

\begin{document}

\title{SETGAN: Scale and Energy Trade-off GANs for Image Applications on Mobile Platforms} 

{\footnotesize 
\thanks{This work was supported in part by the National Science Foundation grants OAC-1910213 , and in part by the U.S. Army Research Office (ARO) grants W911NF-17-1-0485 and W911NF-19-1- 0162.}
}

\vspace{-3.0ex}

\author{Nitthilan Kanappan Jayakodi, Janardhan Rao Doppa, Partha Pratim Pande}
\affiliation{
\institution{School of EECS, Washington State University, Pullman, WA, USA} }
\email{n.kannappanjayakodi@wsu.edu, jana.doppa@wsu.edu, pande@wsu.edu}

\begin{abstract}

We consider the task of photo-realistic unconditional image generation (generate high quality, diverse samples that carry the same visual content as the image) on mobile platforms using Generative Adversarial Networks (GANs).
In this paper, we propose a novel approach to trade-off image generation accuracy of a GAN for the energy consumed (compute) at run-time called Scale-Energy Tradeoff GAN (SETGAN). GANs usually take a long time to train and consume a huge memory hence making it difficult to run on edge devices. The key idea behind SETGAN  for an image generation task is for a given input image, we train a GAN on a remote server and use the trained model on edge devices. We use SinGAN, a single image unconditional generative model, that contains a pyramid of fully convolutional GANs, each responsible for learning the patch distribution at a different scale of the image. During the training process, we determine the optimal number of scales for a given input image and the energy constraint from target edge device. Results show that with the SETGAN's unique client-server based architecture, we were able to achieve 56\% gain in energy for a loss of 3\% to 12\% SSIM accuracy. Also, with the parallel multi-scale training, we obtain around 4x gain in training time on the server.

\end{abstract}



\maketitle

\vspace{-1ex}

\section{Introduction}

Many real-world applications including image/video compression, super-resolution, and virtual reality (VR) are enabled by running deep neural networks (DNNs) on mobile devices. All these applications depend on the fundamental generative task of photo-realistic image synthesis. This involves the generation of high-quality images with the same characteristics of the images employed for training the deep models. A recent generative deep learning framework, namely, {\em Generative Adversarial Network (GAN)} \cite{goodfellow_gan}, is the state-of-the-art approach to generate photo-realistic synthetic images. 
Since the development of GAN framework, GANs have been used for \textcolor{black}{producing photo-realistic images which forms the basis for}  diverse image manipulation tasks including interactive image editing \cite{imageediting1, imageediting2, imageediting3, imageediting4}
, sketch2image \cite{ref_paint2img}, and other image-to-image \cite{pix2pix2016} \cite{ref_imagemani} translation tasks as shown in Fig. \ref{fig:res_paint2image}, \ref{fig:res_super_resolution}. However, GANs employed for producing photo-realistic images are large and complex requiring a significant amount of resources (computational and power) to perform repeated inferences. Since mobile platforms are constrained by resources (power, computation, and memory), there is a great need for low-overhead solutions for these emerging GANs to perform energy-constrained inference. {\em To the best of our knowledge, this is the first work on studying methods to 
deploy emerging GANs to predict complex structured outputs on mobile platforms.} 

There are two unique challenges for the efficient deployment of large GAN models to solve the task of photo-realistic image synthesis. First, {\em multi-image GANs} trained offline on a class of images (e.g., faces) won't work well for a different class of images (e.g., natural scenes). Re-training multi-image GAN models based on one or more images is a possibility, but it is not practical to execute on the edge device due to resource constraints and long training time. Alternatively, we can re-train  multi-image or single-image GAN models on a server or cloud, but large training time will cause high latency for image applications resulting in bad user experience. Second, there is no existing methodology to produce structured outputs (images in our case) of user-specified quality by performing adaptive inference based on the hardness of input examples to improve resource-efficiency. 

\begin{figure}[!t]
    \centering
	\includegraphics[width=\linewidth]{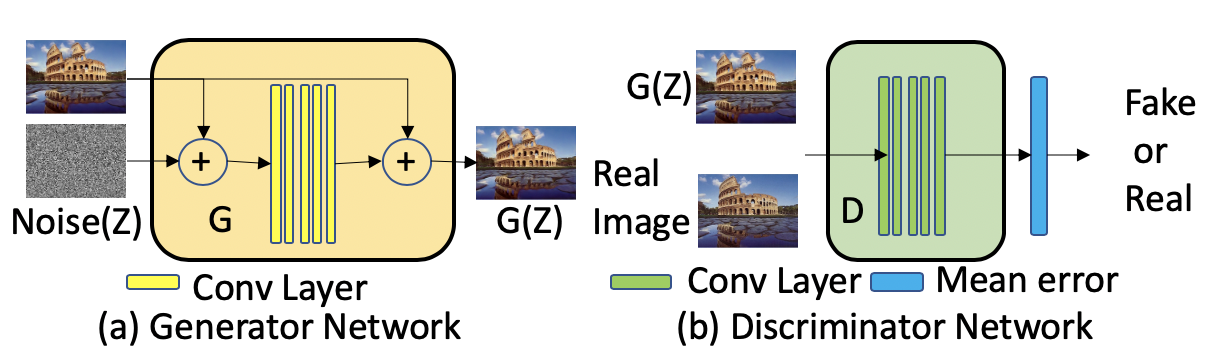}
\vspace{-4ex}

\caption{General structure of a Generative Adversarial Network, where the generator $G$ takes a noise vector $z$ as input and outputs a synthetic sample $G(z)$, and the discriminator $D$ takes the synthetic input $G(z)$ or the real sample $X$ as inputs and predict whether they are real or fake.} 
    \label{fig:gen_dis_arch}
\vspace{-4ex}
\end{figure}

In this paper, we propose a novel approach referred to as {\em SETGAN} to address the above-mentioned two challenges in a principled manner. SETGAN leverages {\em single-image GANs} (smaller model size and faster training/inference time compared to multi-image GANs) for photo-realistic unconditional image generation to automatically trade-off energy consumption and accuracy of inference at {\em runtime} for a target mobile platform. There are two key ideas behind SETGAN. First, SETGAN considers a multi-scale GAN model (GAN models of increasing complexity) that is amenable for embarrassingly parallel training procedures to reduce latency and results in more stable training for high-resolution images.  The main benefit of this multi-scale model is the {\em reuse of computation} from lower scales to perform incremental inference computation at higher scales {\em only if needed}. For any given input image, this multi-scale GAN model is quickly trained on a server or cloud, and the trained model is used for repeated inferences (aka amortization of training cost) to enable image applications, e.g., image editing. Second, given a user-specified accuracy threshold, SETGAN automatically selects the smallest scale GAN that meets the user-threshold to save energy and improve inference time. Simple images can be generated with small-scale GANs to save energy and we only need higher-scale GANs for complex images.


\vspace{0.8ex}

\noindent {\bf Contributions.} The main contributions of this paper include:

\vspace{-1.0ex}

\begin{enumerate}
    \item First study to benchmark GAN for photo-realistic image synthesis applications on mobile platforms.
    \item A novel formalism referred to as SETGAN to trade-off energy and scale of inference using single-image GANs. 
    \item Development of a novel parallel approach for training single-image GANs to achieve 4x speedup over SOTA.
    \item Development of a client-server architecture to effectively apply SETGAN for image editing applications.
    \item Comprehensive experiments on diverse image datasets and applications to show the generality and effectiveness of our approach. Results show that SETGAN achieves 56\% gain in energy 3\% to 12\% in 
    accuracy when compared to the state-of-the-art method called SinGAN networks \cite{singan}.
\end{enumerate}

\vspace{-2ex}
\section{Related Work}



\vspace{0.8ex}

\noindent {\bf Multi-Image GANs.} \textcolor{black}{Initial attempts to apply GANs \cite{goodfellow_gan} suffered from being noisy and incomprehensible. However, growing both the generator and discriminator progressively starting from low-resolution images and adding new layers that introduce higher-resolution details as the training progresses, greatly improved the speed and stability of training leading to successful applications \cite{PGAN, DCGAN}} GANs trained on multiple images have been successfully applied for photo-realistic image generation applications including super-resolution (scaling a low-resolution image to a high-resolution image) \cite{ref_superresolution}, image-to-image translation \cite{pix2pix2016}, and paint2image \cite{ref_paint2img}. However, GANs designed for these applications are very task-specific: models trained for a specific task cannot be used for other tasks without re-training or using different models. Additionally,  these GANs work only for a specific class of image datasets (e.g., faces) for which the GAN has been trained for and fail on other kinds of images (e.g., natural scenes). In contrast, our SETGAN model once trained can be applied to various image manipulation tasks including super-resolution, harmonization, paint2image, and editing. Since we train the model for every new image, our approach is applicable for diverse images and datasets.

\vspace{0.8ex}

\noindent {\bf Single-Image GANs.} Since multi-image GANs are trained using a large number of images from a specific dataset, they try to capture the characteristics of the whole dataset requiring large models making it difficult to run on edge platforms.  This drawback is addressed by GANs which train on single images \cite{singan}. Several recent works fit a deep model to a single training image, but all these methods were designed for specific tasks (e.g., super-resolution \cite{ref_superresolution}). 
Furthermore, unconditional single-image GANs have been explored only in the context of texture generation \cite{texture_syn1, texture_syn2}. In contrast, our SETGAN model is purely generative (i.e., maps noise to image samples) making it suitable for many image manipulation tasks.

\vspace{-2ex}

\subsection{Deploying DNNs on Mobile Platforms}

Prior work to deploy DNNs on resource-constrained mobile platforms is primarily studied for simple classification tasks. 
{\em There is no prior work in this problem space for deploying GANs to predict complex structured outputs}. Existing methods for classification-specific DNNs fall into three categories.

\vspace{0.8ex}

\noindent {\bf Hardware Approaches.} Methods include the design of specialized hardware targeting DNN computations using data-flow knowledge \cite{Eyeriss}, placement of memory closer to computation units \cite{TETRIS}, and on-chip integration of memory and computation \cite{DaDianNao}. Apart from the constraint of application-specific hardware, most of these solutions require analog to digital conversion \cite{Survey_v.Sze}.

\vspace{0.8ex}
\noindent {\bf Software Approaches.} Most important software-level methods are as follows: reduced precision of weights and activations \cite{FPweights}; binary weights 
\cite{BinaryDNN} and its variations like LightNN \cite{QuantizedDNN}; pruning and compression of large DNNs \cite{DBLP:journals/corr/HanPTD15} including energy-aware pruning that was shown to be more energy-efficient \cite{Energy-Aware}; exploiting sparsity of DNN models \cite{Eyeriss}.
There are also approaches that employ a sequence of networks to perform conditional inference (CDL)\cite{CondDeepLearn, C2F:TCAD2018, C2F:TCAD2019, DAC} via Bayesian optimization \cite{MESMO, USEMO} depending on the input example to save computation and energy. Cascaded CNN architecture for face detection \cite{ConvFaceDetect} progressively prune areas of the image that are not likely to contain a face. 
Iterative CNN approach \cite{ICNN-DATE2018} makes multi-stage predictions for images, where a two-stage wavelet transform is applied to produce multiple small-scale images and train separate models to process them progressively to make predictions. Adaptive NN \cite{EnergyConstrainedNN} employs two DNNs --- small one on mobile and the large one on a remote server -- and performs joint optimization 
\textcolor{black}{These methods are for classification tasks and cannot solve tasks with structured outputs.}

\vspace{0.8ex}

\noindent {\bf Hardware and Software Co-Design Approaches.} Many of these approaches involve the design of hardware guided by software level optimization techniques. \cite{FPweights} proposes a compiler-specific to FPGA that analyzes CNNs and generates modules to improve the throughput of the system. In \cite{EIE}, the authors propose the design of the Energy Inference Engine that deploys pruned DNN models. In \cite{Cnvlutin}, the knowledge of compressed sparse-weights guides the hardware to read weights and perform MAC computations only for the nonzero activations, thereby reducing the energy cost by 45\%.  

\vspace{-1ex}

\section{Background and Preliminaries}



\noindent{\bf Generative Adversarial Networks.}
A GAN for unconditional image synthesis involves a generator $G$ which takes a noise vector $z$ as input and outputs a synthetic image that closely resembles the image from a real data set. 
As shown in figure \ref{fig:gen_dis_arch}, a GAN architecture primarily is made of two networks: a) Generator network, and b) Discriminator network. The generator $G$ learns to generate plausible data by taking a random noise vector $z$, and outputs synthetic data $G(z)$; a discriminator $D$ learns to distinguish the generator's fake data $G(z)$ from real data. The discriminator penalizes the generator for producing implausible results. It takes an input $x$ or $G(z)$ and outputs a probability $D(x)$ or $D(G(z))$ to indicate whether it is synthetic or from the true data distribution, as shown in Figure \ref{fig:gen_dis_arch}. 

\vspace{-0.5ex}

\noindent{\bf Training GAN Models.}  The principle to train the generator and discriminator is to form a two-player min-max game, where the generator $G$ tries to generate realistic data to fool the discriminator and discriminator $D$ tries to distinguish between real and fake data. The value function to be optimized is as follows: \textcolor{black}{$\min_{G}\max_{D} V(D,G) = E_{x \sim p_{data}(x)}[log \; (D(x))]+ E_{z \sim p_{z}(z)}[log \; (1-D(G(z)))] $,} where $p_{data}(x)$ denotes the true data distribution and $p_z(z)$ denote the noise distribution. 
When training begins, the generator produces fake data, and the discriminator quickly learns to tell that it's fake. As training progresses, the generator gets closer to producing samples that can fool the discriminator. Finally, if generator training goes well, the discriminator gets worse at telling the difference between real and fake. It starts to classify fake data as real, and its accuracy decreases.

\newcommand\numberthis{\addtocounter{equation}{1}\tag{\theequation}}



\vspace{0.4ex}
\noindent {\em \bf Discriminator:} The discriminator in the GAN model is a classifier. It tries to distinguish real data from fake data created by the generator. 
The discriminator's training data comes from two sources: a) Real data samples, which act as positive examples during training; and b) Fake data samples created by the generator, which act as negative examples during training. The discriminator connects to two loss functions. During discriminator training, the discriminator ignores the generator's loss and just uses the discriminator loss. The discriminator classifies both real data and fake data from the generator. The discriminator's loss penalizes the discriminator for misclassifying a real sample as fake or a fake sample as real. The discriminator updates its weights 
from the discriminator's loss. 

\vspace{0.4ex}

\noindent {\em \bf Generator:} The generator part of a GAN learns to create fake samples 
to make the discriminator classify its output as real. Generator's training requires tighter integration between the generator and the discriminator. The portion of the GAN that trains the generator includes: a) random input; b) generator network, which transforms the random input into a data sample; c) discriminator network, which classifies the generated data as real or fake; and d) discriminator output generator loss, which penalizes the generator 

\vspace{0.4ex}

\noindent{\bf Structural Similarity Index Metric .} For predicting the perceived quality of synthesized images in SETGAN approach, we employ SSIM score that measures the similarity between two images. 
The $SSIM$ between two images $x$ and $y$ is given by equation $SSIM(x,y) = \frac{(2\mu_x\mu_y + C_1) + (2 \sigma _{xy} + C_2)} 
    {(\mu_x^2 + \mu_y^2+C_1) (\sigma_x^2 + \sigma_y^2+C_2)}$
where $\mu_x$, $\mu_y$ are the means and $\sigma_x$, $\sigma_y$ are the variances of images $x$, $y$; and $\sigma_{xy}$ is the covariance of $x$ and $y$.


\vspace{-2ex}

\section{SETGAN Framework}



\noindent {\bf Motivation.} Image editing applications such as Paint2Image and super-resolution are time-sensitive and utilize an image synthesis operation as part of their workflow. 
Deploying large GAN models on mobile platforms requires a lot of resources in terms of computation, energy, and memory.  Multi-image GANs such as PGAN \cite{PGAN} have a) a huge number of parameters, and b) take a long time to train for new image datasets (usually in days). Single-image GANs such as SinGAN \cite{singan} have relatively less number of parameters, but they still take a long time (in order of hours) to train a given input image. 

In the SETGAN framework, we propose a single-image multi-scale GAN model, which {\em automatically} adapts the number of scales based on the complexity of input image in a client-server architecture. The key idea is to use a small number of scales for a simple image (which consists of small structures and textures) while using more number of scales for complex images (which has large objects). We were able to achieve 56\% gain in energy for a loss of 3\%-12\% SSIM accuracy.  Furthermore, we propose a novel training methodology to reduce the training time by a factor of 4X. We adapt the SinGAN model \cite{singan} since it has a relatively small number of parameters making it an ideal candidate for image editing applications in a client-server architecture. In the following section, we describe our solution in detail. First, we describe the SETGAN model including generator, discriminator, training procedure, and inference methodology. Second, we discuss the client-Server architecture including the different modes of operation of the GAN model for image editing applications. Finally, we instantiate our SETGAN framework for diverse image editing application workflow.


\begin{figure}[h]
    \centering
\includegraphics[width=\linewidth]{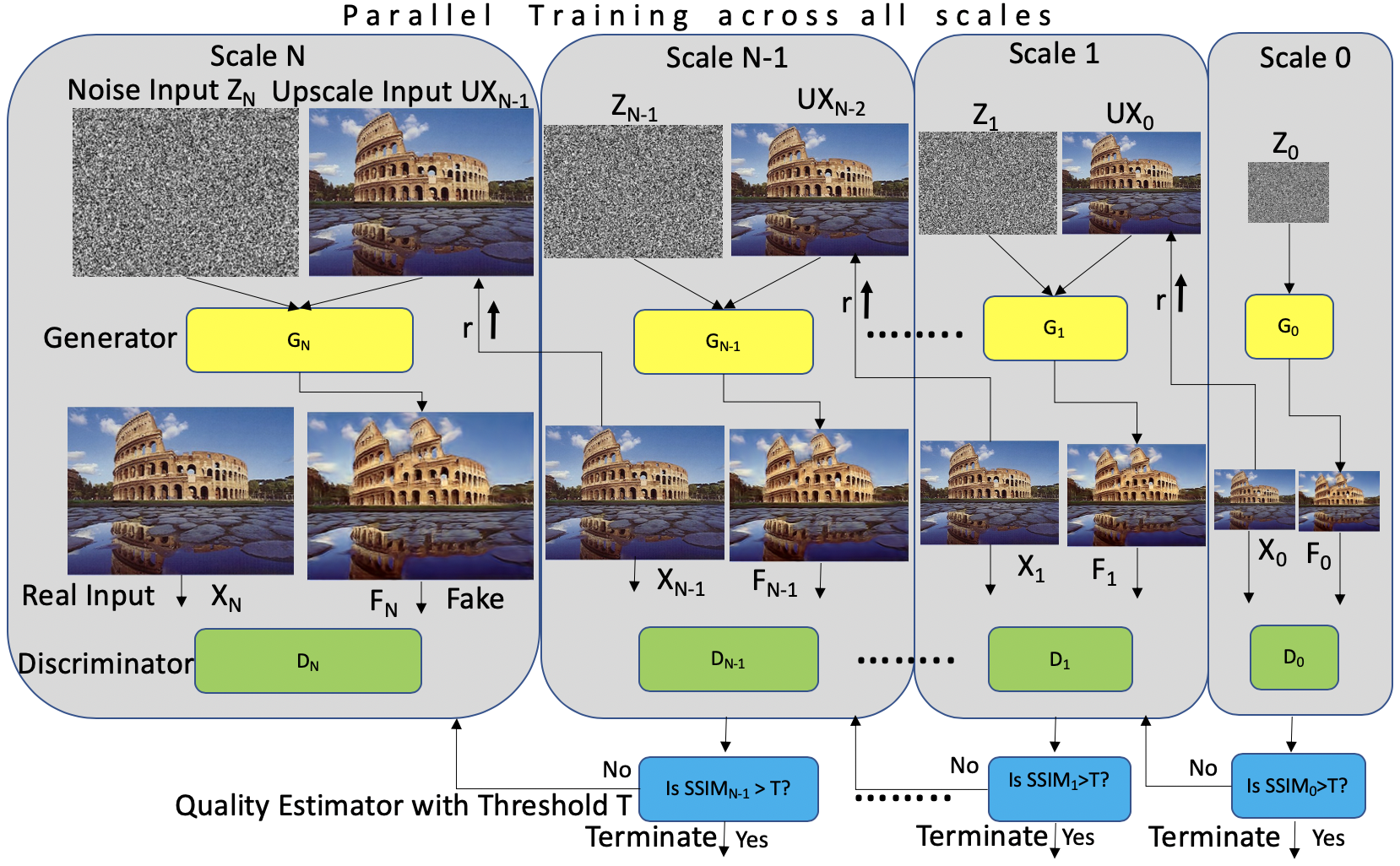}
\vspace{-4ex}

\caption{Illustration of SETGAN approach with multi-scale GAN model. An abstract model with $N$ scales from coarsest-scale GAN $0$ to finest-scale GAN $N$. Each scale consists of a generator $G_i$, giscriminator $D_i$, and a quality estimator $SSIM_i$ to decide when to terminate. Generator $G_i$ takes a noise vector $Z_i$ and up-scaled real input $X_{i-1}$ and generates a fake image $F_i$. Discriminator $D_i$ takes $F_i$ or real input $X_i$ as input and classifies them as fake or real. $SSIM_i$ checks whether SSIM between the generated fake image $F_i$ and the original input image $X$ is greater than the user-specified threshold $T$.} 
\vspace{-4ex}

\label{fig:setgan_arch}
\end{figure}
\vspace{-0.50ex}

\subsection{SETGAN Model, Training, and Inference}

\noindent {\bf Multi-scale SETGAN Model.} Our model tries to capture the internal statistics of the given single-image $X$ for training. The unconditional image generation task is conceptually similar to the conventional GAN setting. However, instead of the whole image, we train the model using patches of a single image. As shown in the Figure \ref{fig:setgan_arch}, we create a pyramid of images from the given input $X$: $X_0,\cdots,X_N$, where $X_i$ is a down-sampled version of $X$ ($X_0$ being the coarsest image) by a factor $r^{N-i}$, for some $r > 1$. The multi-scale SETGAN model consists of a list of generators for each down-sampled version of $X$, $G_0,\cdots,G_N$, a associated list of discriminators $D_0,\cdots,D_N$ and Quality estimators $SSIM_0,\cdots,SSIM_{N-1}$.

\vspace{0.5ex}

\noindent{\bf Parallel Training Algorithm.} Image-editing applications are time-sensitive. However, the SinGAN model based on which we have developed our SETGAN model employs a serial training procedure, which is very expensive. Starting from the coarsest-scale, SinGAN uses the fake image $F_{i-1}$ generated at the end of training of the previous stage as input for the next generator $G_i$'s input. {\em In contrast, in SETGAN, we employ the scaled (real) input image $X_{i-1}$, which allows us to perform embarrassingly parallel training of models at all scales}. The key reason we are able to do this is as follows: $F_{i-1}$ is an approximation of $X_{i-1}$ and this approximation helps us in achieving 4x reduction in training time \textcolor{black}{with negligible loss in quality of generated images. Since the goal of image editing tasks is for the generated image to be similar and not exact to the original input image, this negligible loss in quality is not noticeable.} 

\vspace{-1ex}
\begin{figure}[h]
    \centering
	\includegraphics[ width=\linewidth]{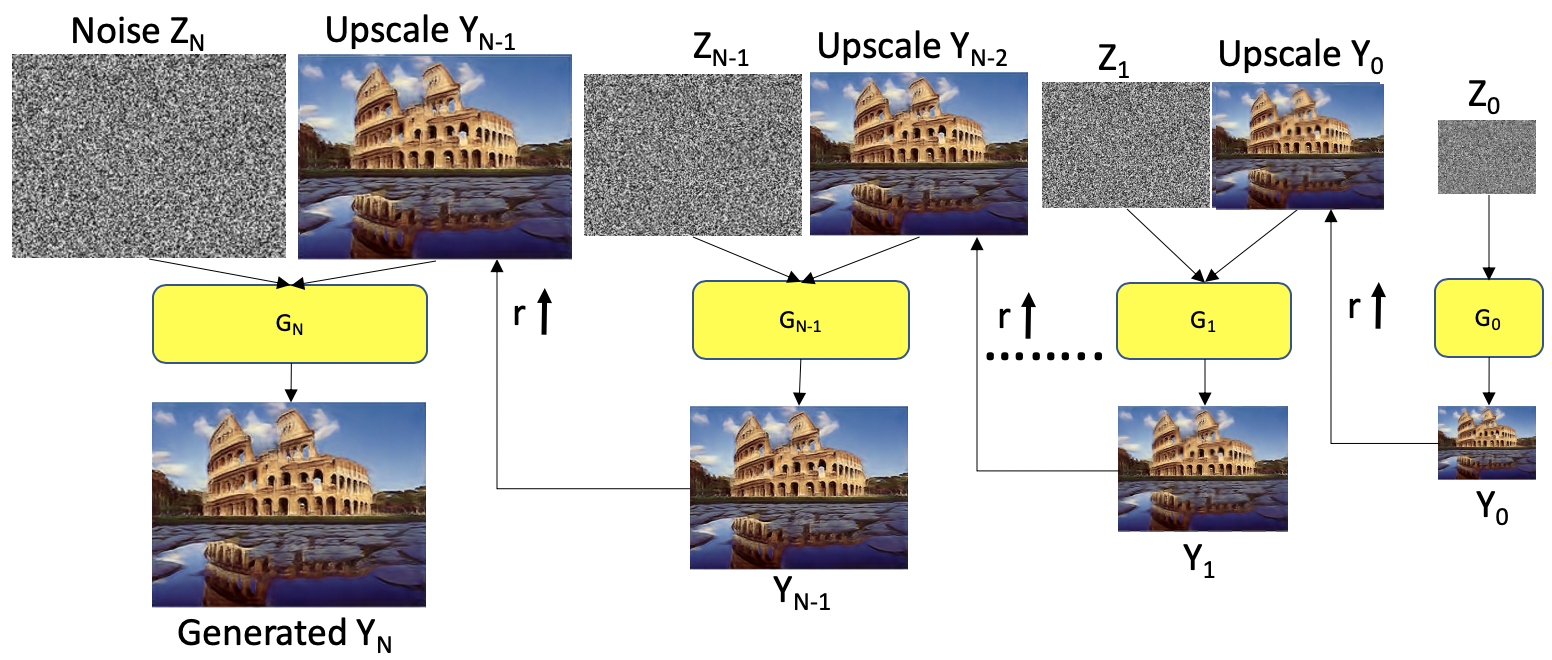}
\vspace{-2ex}

\caption{SETGAN inference: The unconditional image synthesis starts at coarsest scale $G_0$ taking noise vector $Z_0$ as input and passes through all generators upto finest scale $G_N$.} 
    \label{fig:infer_archtecture}
\vspace{-2ex}

\end{figure}
\vspace{-1ex}
\noindent {\em Training Procedure.} Given an input image $X$ and a target SSIM threshold for early exit $T$, we perform the training of models at all $N$ scales in parallel. Each scale $i$ is made of a generator $G_i$, discriminator $D_i$, and a quality estimator $SSIM_i$. They take as input 1) $X_i$, the image the generator has to match for this scale $i$; 2) Up-scaled $X_{i-1}$, the image from a coarser scale $X_{i-1}$ up-scaled by a factor $r$; 3) Gaussian noise vector $Z_n$. Each generator $G_i$ is responsible for producing realistic image samples w.r.t the patch distribution in the corresponding image $X_i$. As described in the background section, the generator achieves this goal via adversarial training, where $G_i$ learns to fool an associated discriminator $D_i$, which attempts to distinguish patches in the generated samples from patches in $X_i$. At the end of adversarial training, quality estimator $SSIM_i$ checks whether SSIM between the generated fake image $F_i$ and the original input image $X$ is greater than the early exit threshold $T$ or not. If this condition is satisfied, then we terminate the training procedure at this scale and use the models from $0, 1,\cdots,i$ for our inference. 

\vspace{0.5ex}

 {\bf a) Generator:} The generator at scale 0 is purely generative, i.e., $G_0$ maps Gaussian noise $Z_0$ to an image sample $F_0$. However, in all other scales, $G_i$ maps the Gaussian noise $Z_i$ to produce the incremental image from $X_{i-1}$ to that at $X_i$. All the generators have a similar architecture as shown in figure \ref{fig:gen_dis_arch}. Each generator is a fully-convolutional network with 5 convolution blocks. Each block is made of three layers namely (3x3) Conv layer, BatchNorm followed by a ReLU, except for the last block whose activation is Tanh. It starts with 32 kernels per block at the coarsest-scale and increases by a factor of 2 for every 4 scales. The noise vector is added to the scaled image prior to being fed to the convolution blocks, ensuring that the GAN does not disregard the noise. 

\vspace{0.5ex}

{\bf b) Discriminator:} Each generator $G_i$ works with a corresponding Markovian discriminator $D_i$ that classifies the overlapping input patches as real or fake. A WGAN-GP loss \cite{wp_loss_gan} is used to increase the training stability. 
The architecture of $D_i$ is similar to that of the generator as shown in figure \ref{fig:gen_dis_arch}. However, the last block at each scale includes neither normalization nor activation.  It is followed by a mean error block which averages the error over entire image. 
\vspace{-4ex}

\begin{align*} 
\min_{G_n}\max_{D_n} L_{adv}(D_n,G_n) + \alpha*L_{rec}(G_n) \numberthis \label{eq:setgan_loss}
\end{align*}
\vspace{-3ex}

Equation \ref{eq:setgan_loss} describes the training loss of the $i$th scale GAN model. $L_{adv}$ represents the adversarial loss. 

\vspace{0.8ex}

\noindent{\bf Inference Algorithm.} As shown in figure \ref{fig:infer_archtecture}, the image generation starts at the coarsest-scale generator $G_0$ which takes input as noise vector $Z_0$ and produces a generated image $Y_0$. At each scale $i>0$, generator $G_i$ takes the image generated at a coarser scale $Y_{i-1}$ and noise vector $Z_i$ as input and produces a finer image $Y_i$. This procedure passes through all the generators upto the scale determined by the training algorithm to meet the quality threshold. However, as shown in figure \ref{fig:infer_archtecture}, during the inference stage, we do not use the real images. Because the generators are fully convolutional, we can generate images of arbitrary size and aspect ratio at test time (by changing the dimensions of the noise maps).

\begin{figure}[h]
    \centering
	\includegraphics[ width=0.80\linewidth]{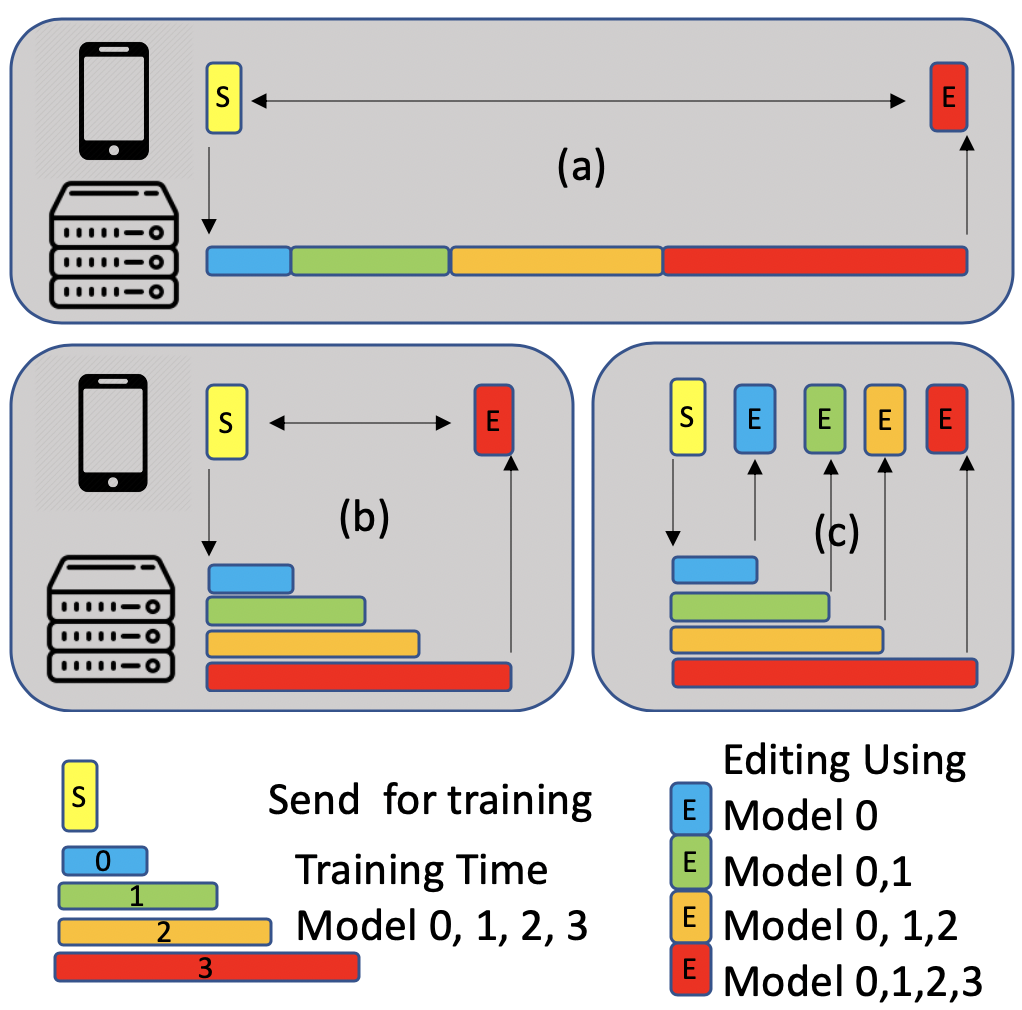}
\caption{Client-server architecture: Describes the three modes of operation for the SETGAN Framework (a) SinGAN Baseline, (b) Parallel training, and (c) Progressive update.} 
    \label{fig:client_server_arch}
\vspace{-5ex}

\end{figure}

\subsection{Client-Server Architecture}

The training procedure of GAN models involve thousands of iterations of forward and backward propagation steps, which are very compute-intensive and are usually executed using a GPU. 
\textcolor{black}{Hence, training GANs on mobile platforms is impractical. Furthermore, any image manipulation task involves repeated editing of the given input image until the user is satisfied (i.e., repeated inference calls). 
The {\bf Naive client-server architecture} involves sending the image to be edited to the server and for every image edit, the server sends the edited image back to the client: {\em all training and inference calls are executed on the server}. This causes network latency to the user after every edit and increases the client-server memory bandwidth for every edit operation.}
\textcolor{black}{Hence, we propose a modified client-server architecture to efficiently use SETGAN for image-editing applications and discuss three different modes of operation.}

\vspace{0.7ex}

\noindent{\bf [\textcolor{black}{Mode} 1] SinGAN Baseline:} As shown in figure \ref{fig:client_server_arch} (a),  we employ the existing SinGAN model for our training. This is our baseline mode. In SinGAN, the training happens sequentially from the coarsest scale $G_0$ to the finest scale $G_N$ for any given input image. Once the training is complete, all the models are sent together in a one-shot to the edge device.\textcolor{black}{Unlike the default mode, since the entire model is available on the mobile device, repeated inference happens without the network latency and does not increase the client-server memory bandwidth for every edit operation}

\vspace{0.7ex}

\noindent{\bf [\textcolor{black}{Mode} 2] Parallel Training:} As shown in figure \ref{fig:client_server_arch} (b), in this mode of operation, we employ SETGAN model for our training. 
Since the SETGAN models at all the scales are trained in parallel and the {\em best} scale to meet the quality threshold is determined automatically, it takes significantly less time to train SETGAN for a given input image. Once training is complete, all the models from $G_0$ to $G_{best}$ (best $\leq N$) are sent in a one-shot to the edge device.

\vspace{0.7ex}

\noindent{\bf [\textcolor{black}{Mode} 3] Progressive Update:} As shown in figure \ref{fig:client_server_arch} (c), in this mode of operation, we use SETGAN model for our training similar to mode \#2. However, we send the trained model to the edge device as and when it's training is complete. Since we do not wait for the full training to be complete, we can start using our image-editing application with the available models which provide a coarser output. The generated images would be refined as and when finer-scale models are available. This mode reduces the time to start the editing application and results in a better user experience.

\vspace{-3ex}

\subsection{SETGAN for Image Applications}

In this section, we discuss the use of the SETGAN framework for various image-editing applications. Recall that a trained SETGAN generator model is capable of producing images with the same patch distribution as that of the single training image.  Different image manipulation tasks can be performed by injecting (a down-sampled version of) an image into the pyramid of generators at some scale $i \leq  N$. When this image passes through the various generators, it produces an output that matches the patch distribution to that of the training image. We can perform diverse image manipulations by using an appropriate scale at which we introduce the image.

\vspace{0.7ex}

\noindent {\bf Key Steps.} On receiving an input image for editing, the following sequence of steps are performed: 1) The image is sent to the server; 2) Using the input image, the SETGAN model is trained for the optimal number of scales (say {\em best});  3) The trained SETGAN generator models ($G_0,\cdots,G_{best}$) are sent to the edge device; and 4) Using the input image and the received SETGAN model, we can repeatedly perform various image-editing operations on the edge device. We instantiate SETGAN for three diverse image-editing applications that will be employed in our experimental evaluation.



\vspace{0.7ex}

\noindent{\bf Super-resolution:} Given a low-resolution (LR) input image, we want to scale it up by a factor $s$ to produce a high-resolution image as shown in Fig~\ref{fig:res_super_resolution}. 
To support a scale factor $s$, we train SETGAN for $N$ scales such that $r = \sqrt[\uproot{2}k]{s}$. 
During inference, we up-sample the LR image by a factor $r$. This along with the noise vector is fed to the last generator $G_N$. We repeat $k$ times to obtain the final high-resolution output. 


\vspace{0.7ex}

\noindent{\bf Paint2Image:}  Given a clip-art image as input, we want to generate a photo-realistic image as output (illustrated in Fig~\ref{fig:res_paint2image}). 
The clip-art image is an LR representation of the output image we need to generate. Hence, we downsample the clip-art image and feed it to one of the coarse scales like $G_1$ or $G_2$. This preserves the global structure of the painting while texture and high-frequency information matching the original image generates realistic photos. 



\vspace{0.7ex}

\noindent{\bf Harmonization:} As shown in Fig~\ref{fig:res_harmonization}, the Eiffel tower object is pasted onto to image which looks like a painting. Harmonization is the process of realistically blending this pasted object with the background image. In this case, we inject a down-sampled version of the composite image at the inference stage. Unlike paint2image that works from the coarse scales, we inject the image at the finer scales of the generator in this application. Scales $G_{N-1}, G_{N-2}, G_{N-3}$ generally provide a good balance between preserving the object's structure and transferring the background texture.

\vspace{-1ex}

\section{Experimental Setup}




\noindent {\bf Hardware setup.} All our experiments were performed by deploying SETGAN on an \texttt{ODROID-XU4} board, \cite{odroidboard}. 
We employ SmartPower2 \cite{SP2} to measure power. 


{\noindent \bf GAN training on server.} We set the minimal dimension at the coarsest-scale to 25px, and choose the number of scales $N$ such that the scaling factor $r$ is as close as possible to 4/3. For all the results (unless mentioned otherwise), we re-sized the training image to the maximal dimension 256 px. Thus, the maximum number of scales $N$ is 9. At each scale, the weights of the generator and discriminator are initialized randomly. We train each scale for 2000 iterations. In each iteration, we alternate between three gradient steps for the generator and three gradient steps for the discriminator. We employ the Adam optimizer with a learning rate of 0.0005 (decreased by a factor of 0.1 after 1600 iterations), and momentum parameters $\beta_1$= 0.5 and $\beta_2$= 0.999. The weight of the reconstruction loss $\alpha$ is set to 10, and the weight of the gradient penalty in the WGAN loss is set to 0.1. All Leaky-ReLU activations have a slope of 0.2 for negative values. The training uses an RTX 2080Ti GPU. 


\noindent {\bf Datasets.} 
We employed the Berkeley Segmentation Database (BSD) \cite{BSD} and Places \cite{places_dataset} datasets \textcolor{black}{consisting of 500 and 10 million images respectively. Our approach works with one image at a time.}

\vspace{-0.5ex}

\begin{table}[h!]
\centering
 \begin{tabular}{||l r r||} 
 \hline
 GAN Type & Memory (MB)  & Training Time (Hours) \\ [0.5ex] 
 \hline\hline
 PGAN & 3343 &  24.00 \\ 
 SinGAN & 98 &  1.00 \\
 SETGAN & 98 &  0.25 \\ 
 \hline
 \end{tabular}
 \caption{Training time and memory required for optimized SETGAN and state-of-the-art GAN models.} 
    \label{fig:single_vs_multiple}
\vspace{-5.0ex}

\end{table}

\noindent {\bf Energy objective.} Energy consumption of SETGAN is measured in terms of the normalized energy-delay product (EDP). EDP = $\sum$P $\Delta$t $\cdot T$, where $\Delta$t is the time interval at which we record the power $P$ and $T$ is the total execution time. As power is measured at a regular interval, we simply calculate EDP as EDP = $P_{avg}$ $\cdot$ $T^{2}$ . This value is normalized with EDP of SETGAN with all scales (0 to $N$). 

\vspace{0.5ex}

\noindent {\bf Accuracy objective.} The accuracy of the unsupervised photo-realistic image generation from the SETGAN method is measured using the Structural Similarity Index Metric (SSIM) score \textcolor{black}{which measures the quality of the generated images similar to how human perceive images unlike metrics such as Peak Signal-to-Noise Ratio.}  





\vspace{-1.0ex}
\section{Results and Discussion}

In this section, we compare the performance of SETGAN with state-of-the-art GAN models to demonstrate its effectiveness. 

\begin{figure}[h]
    \centering
	\includegraphics[trim={20 20 45 40},clip, width=0.495\linewidth]{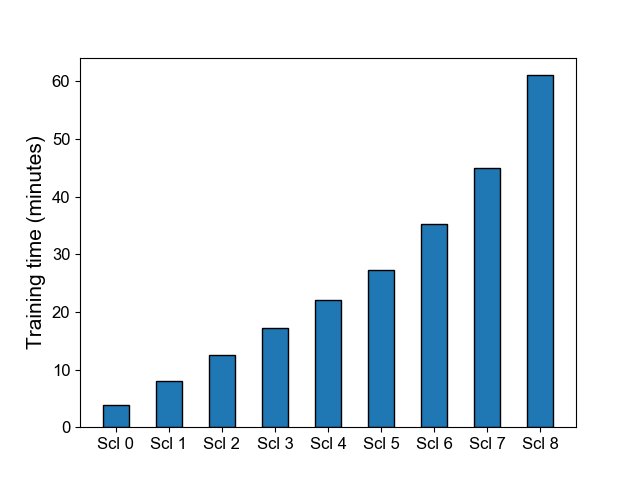}
	\includegraphics[trim={20 20 45 40},clip, width=0.495\linewidth]{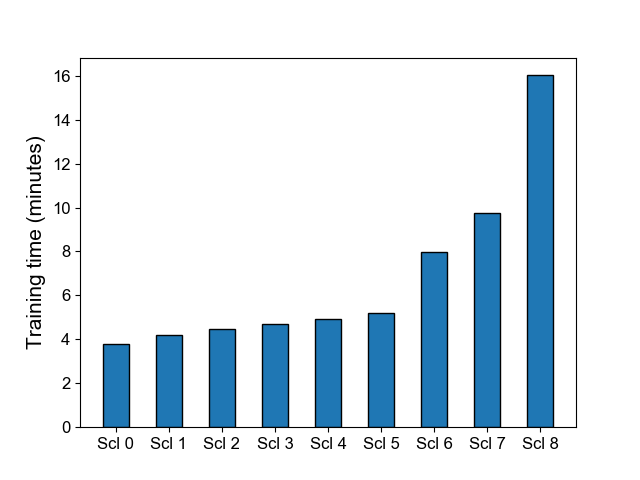}

	\vspace{-3ex}
\caption{Effect of training time with the number of scales for a single image GAN: (Left) Baseline SinGAN training time across scales and (Right) Optimized SETGAN training 
} 
	\vspace{-3ex}

    \label{fig:train_time_perf}
\end{figure}
\vspace{-1.0ex}

\subsection{SETGAN vs. PGAN and SinGAN}

We compare the memory and training time of SETGAN in comparison to state-of-the-art GAN models: PGAN, a multi-image GAN \cite{PGAN}, and SinGAN, a single-image GAN \cite{singan}. 

\vspace{0.8ex}

\noindent {\bf Results for memory and training time.} PGAN uses multi-scale GAN models and is trained on specific datasets with thousands of images. SinGAN performs serial training using a single image. Table~\ref{fig:single_vs_multiple} shows the results comparing SETGAN with PGAN and SinGAN. SETGAN and SinGAN models consume 34x less memory than the PGAN model. This is expected because PGAN needs to store the information regarding different kinds of images used for training, whereas SETGAN and SinGAN \textcolor{black}{architectures are intended to be trained only on a single image.} \textcolor{black}{Furthermore, the SETGAN memory can be further optimized with progressive update mode as discussed later.} The average training times of PGAN, SinGAN, and SETGAN are 24 hours, 1 hour, and 15 minutes respectively. \textcolor{black}{Thus, PGAN and SinGAN are not suitable for time-sensitive image editing applications.} The fast training time of SETGAN is due to our proposed parallel training.

\begin{figure}[h]
    \centering

	\includegraphics[trim={10 20 45 40},clip, width=0.495\linewidth]{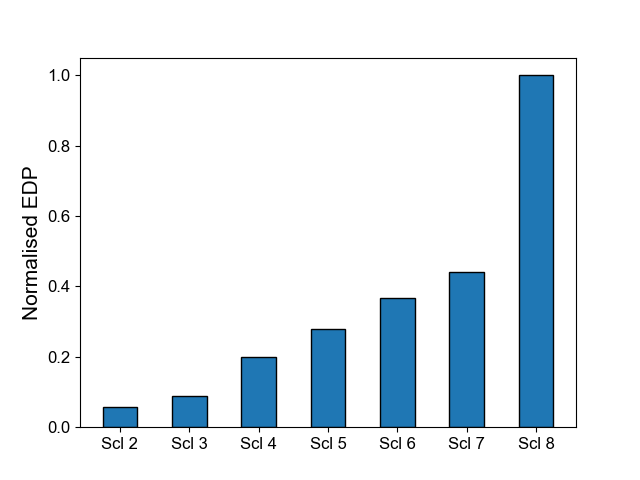}
	\includegraphics[trim={10 20 45 40},clip, width=0.495\linewidth]{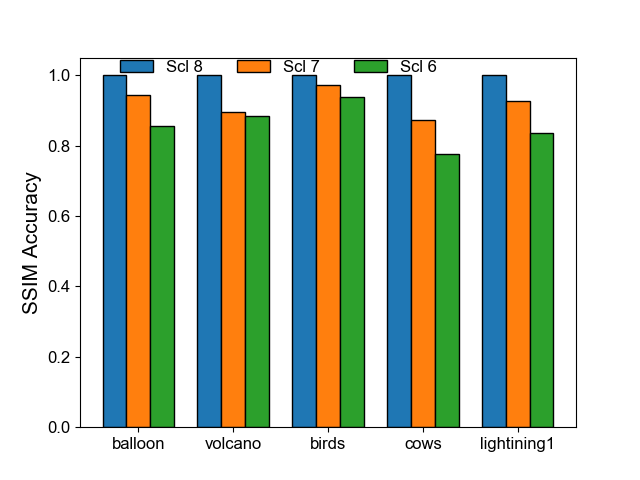}
	\vspace{-2ex}
\caption{Energy and accuracy trade-off of SETGAN for unconditional photo-realistic image synthesis on mobile platform: (Left) Normalised EDP consumed at different scales of SETGAN generator network and (Right) SSIM Accuracy of different images across scales 6, 7, and 8.} 
\vspace{-2.0ex}

    \label{fig:acc_vs_edp}
\end{figure}

\noindent{\bf Effect of parallel training.} In Fig. \ref{fig:train_time_perf} (left), we show the time required to train the baseline SinGAN model of scales 0 to 8. We can see that training time increases gradually with the number of scales (4 mins for the coarsest scale and 60 mins to train for all scales) due to serial training. On the other hand, due to parallel training, in fig. \ref{fig:train_time_perf} (right) SETGAN  can train all nine scales in 15 minutes resulting in 4X improvement over SinGAN. All these results demonstrate that SETGAN is much better suited for image-editing applications 

\vspace{-2.0ex}
\begin{figure}[h]
\centering
\begin{minipage}[b]{.48\linewidth}
	\includegraphics[trim={10 20 45 40},clip, width=\linewidth]{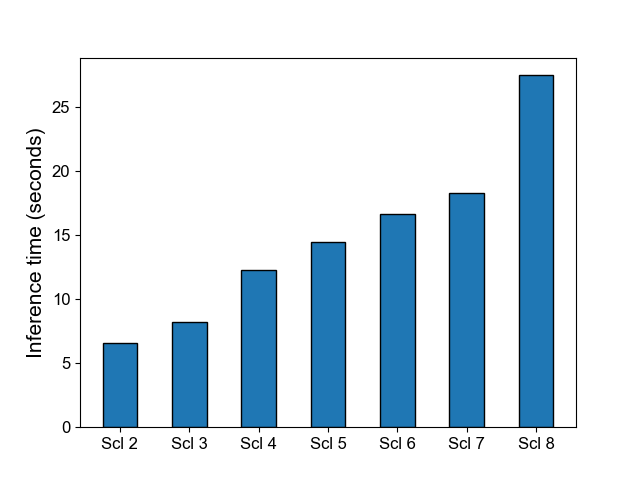}
\vspace{-2.0ex}
\caption{Inference time with SETGAN for photo-realistic image synthesis on mobile platforms: Variation of inference time from scales 2 to 8.}
\vspace{-3.0ex}

    \label{fig:inf_time_perf}
\end{minipage}
\hfill
\begin{minipage}[b]{.48\linewidth}
		\includegraphics[trim={10 20 45 40},clip, width=\linewidth]{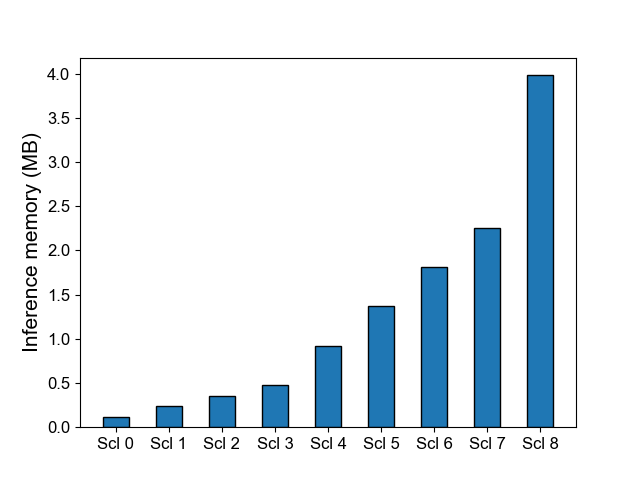}
\vspace{-2.0ex}

\caption{Performance of SETGAN with client-server architecture 
Variation of inference model memory 
across scales from 0 to 8.} 
\vspace{-3.0ex}

    \label{fig:inf_memory}
\end{minipage}
\vspace{-2.0ex}

\end{figure}
\subsection{SETGAN Inference across Scales} 

The SETGAN model trained on the server is sent to the client (mobile device). In this section, we discuss the results of SETGAN inference on mobile platforms. 



\vspace{0.7ex}

\noindent{\bf Accuracy of Generator vs. Normalised EDP.} As shown in Fig. \ref{fig:acc_vs_edp}, we compare the accuracy and energy consumed by the generator network of SETGAN at different scales. Recall that we can control the overall compute of SETGAN using the quality estimate threshold of $T$. By varying this threshold, we can exit early reducing the number of scales used by the image-editing application. We compare the normalized EDP and SSIM accuracy trade-off by a varying threshold of $T$. In figure \ref{fig:acc_vs_edp}(left), we plot the normalized EDP consumed for different scales by averaging across all images in the dataset. We observe that there is a 56\% reduction in energy from scale 8 to scale 7.  For the same reduction in energy, in fig. \ref{fig:acc_vs_edp} (right), we see a drop of 6\%, 10\%, 3\%, 13\%, 7\% in SSIM accuracy for different images such as balloon, volcano, birds, cows, and lightning. Similarly, the energy drops by another 8\% form scale 7 to 6 for a drop of 10\%, 1\%, 4\%, 9\%, 8\% in SSIM accuracy respectively. We see that for simple images such as balloons and birds (contains smaller structures), we see a very small drop in accuracy across scales, thereby use a fewer number of scales and consumes 56\% lesser energy. GANs are generally evaluated for the capability to generate varied images and finding good metrics is still a open problem within ML community. Hence, qualitative results in the form of generated images are shown. Although we use SSIM to measure the closeness with the input image, the drop in SSIM accuracy does not reflect the true performance of GANs. Therefore, similar to prior work, we will present qualitative results for our SETGAN in subsequent sections as user-experience is more important for image-editing applications. 

\vspace{0.7ex}

\noindent {\bf Inference Time.} In this part, we discuss the time taken for running the SETGAN generator model across different scales. As shown in fig \ref{fig:inf_time_perf}, we see that there is a 30\% drop in inference time from scale 8 to scale 7. We also see another 10\% drop in inference time from scale 7 to scale 6. From these results, we make the observation that by reducing the number of scales of a generator, we will be able to get faster feedback for different image-editing applications. 

\subsection{Performance of Client-Server Architecture}

In this section, we study the properties of the SETGAN model which enable the different modes of the client-server architecture. The latency for the client-server architecture depends on the memory of the SETGAN generator model across different scales sent from the server to the mobile device. We discuss the impact of memory size of the model for the different modes of the SETGAN architecture.

\begin{figure}[h]
    \centering
	\includegraphics[trim={20 20 45 40},clip, width=0.495\linewidth]{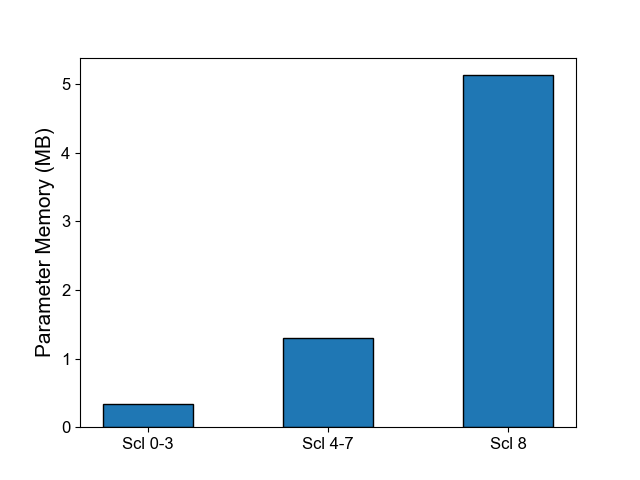}
	\includegraphics[trim={10 20 45 40},clip, width=0.495\linewidth]{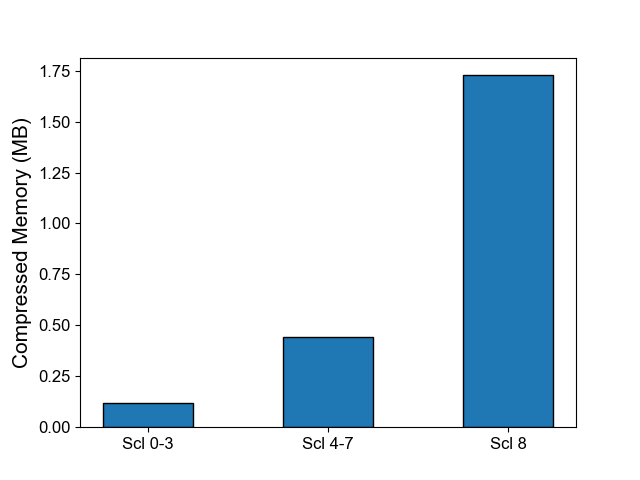}
\vspace{-3.0ex}

\caption{Memory consumed by SETGAN generator at each scale: (Left) Memory size of parameters and (Right) Compressed memory size.} 
\vspace{-4.0ex}

    \label{fig:train_perf}
\end{figure}

\vspace{0.8ex}

\noindent{\bf Parallel training (Mode \#2). } In this mode, after training the SETGAN model, the entire model is sent to the mobile device. In Fig. \ref{fig:inf_memory}, we show the inference memory required for the optimal number of scales of the SETGAN generator model. The generator for the finest-scale (8) takes 4MB of memory. The latency to transfer a 4MB model is minuscule when compared to the training time. From the Fig. \ref{fig:inf_memory}, we see a 44\% decrease in memory of the SETGAN generator model from scale 8 to scale 7 and another 19\% decrease in memory when we go from scale 7 to scale 6 

\vspace{0.8ex}

\noindent{\bf Progressive update (Mode \#3). } is a mode where we use the image-editing application as and when a model is available for use. The latency also depends on the model size. In fig. \ref{fig:train_perf} (Left), we show the memory consumed by the SETGAN generator at each scale. Generator model at scale 0-3 consumes the smallest memory across all scales (0-8). This is because the first four scales of the network consist of 32 filters. The number of filters doubles for every four scales, i.e., scales 4-7 consists of 64 filters and scale 8 consists of 128 filters. This explains the memory consumed at the respective scales. Also, the compressed memory consumed at each scale reduces the memory size by a factor of 3x as shown in Fig. \ref{fig:train_perf} (Right).  
The latency for a model to be used on the mobile device depends on the training time as shown in figure \ref{fig:train_time_perf} and the time it takes to transfer the trained model to the mobile. For the first four scales, the models will be trained at around the same time of 4-5 minutes and the size of the model is 120KB each, which could be transferred in negligible time. 
For a progressive update mode of operation,  we could start using the generator model from 8 minutes at scale 6, and successively use the scale 7 model at 10 minutes and use all the 8 scales at 16 minutes. Thus, the image-editing application could be improved adaptively to further reduce the overall usage performance by another 50\%.
\textcolor{black}{Hence, the overall running time from ``send for training'' to ``Editing'' depends on three major components: a) The Training time ranges between 4-16 minutes based on scale; b) The model transfer time which is negligible due to small model and image size; and c) The Inference time (from the previous section) of around 15-25 seconds for each edit operation. However, the user interaction depends only on the inference time since the user starts editing only after the model is received on the edge device.}
\vspace{-1.0ex}

\subsection{Image-Editing Applications}

We now discuss the quantitative and qualitative results of SETGAN for four diverse image-editing applications. 

\begin{figure}[h]
    \centering
	\includegraphics[ width=\linewidth, height=0.5\linewidth]{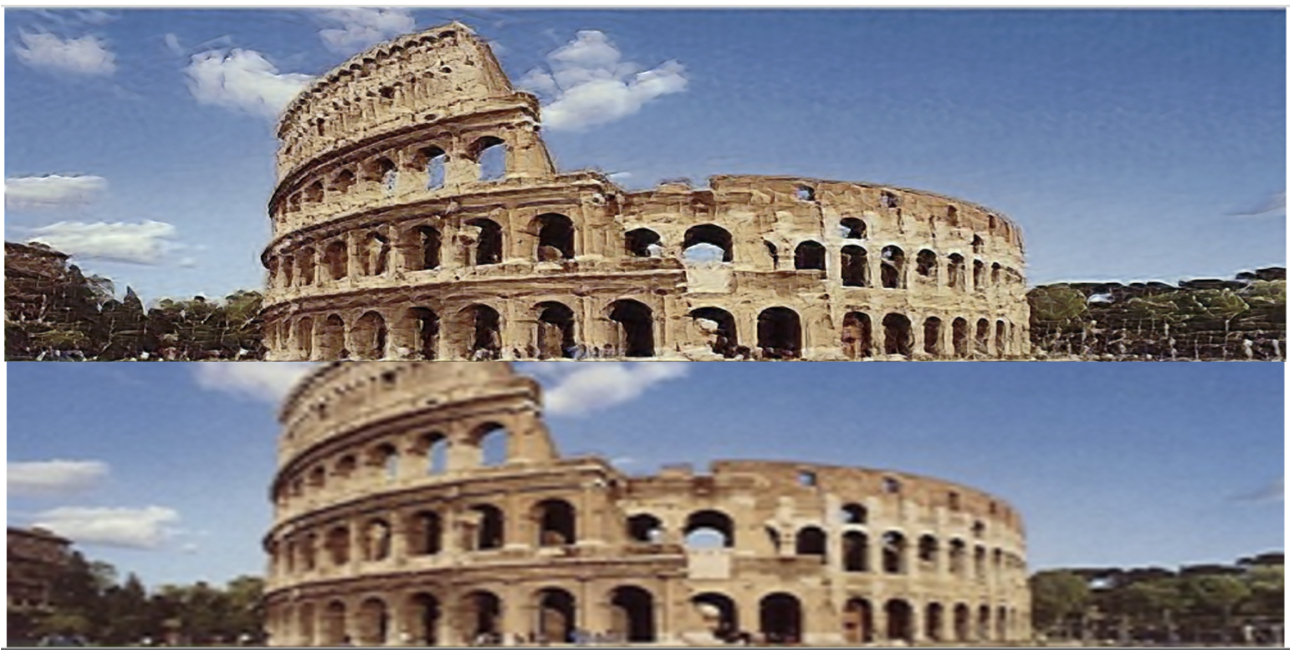}
	\vspace{-4ex}
\caption{Super-resolution: Colloseum image scaled by a factor of 4x generated using (top) SETGAN model and (bottom) Bilinear interpolation.} 
\vspace{-4.0ex}

    \label{fig:res_super_resolution}
\end{figure}

\vspace{0.8ex}

\noindent{\bf Super-resolution.} We train a Colosseum image using the SETGAN model for super-resolution applications. Using the trained SETGAN model, we would be able to scale the low-resolution image for different scales such as 2x, 4x, 8x, etc. In Fig. \ref{fig:res_super_resolution}, we scale the 256x256 image by 4x factor to produce 1024x1024 output images using SETGAN model (top) and bilinear interpolation (bottom). Since the bilinear interpolation is not a generative model, the output image is blurred. However, the image generated using SETGAN has sharp and crisp features since it uses the features at lower-resolution to generate features at higher-resolution. The same effect is seen even at higher scale factors of 8x and 16x.

\begin{figure}[h]
    \centering
	\includegraphics[ width=\linewidth]{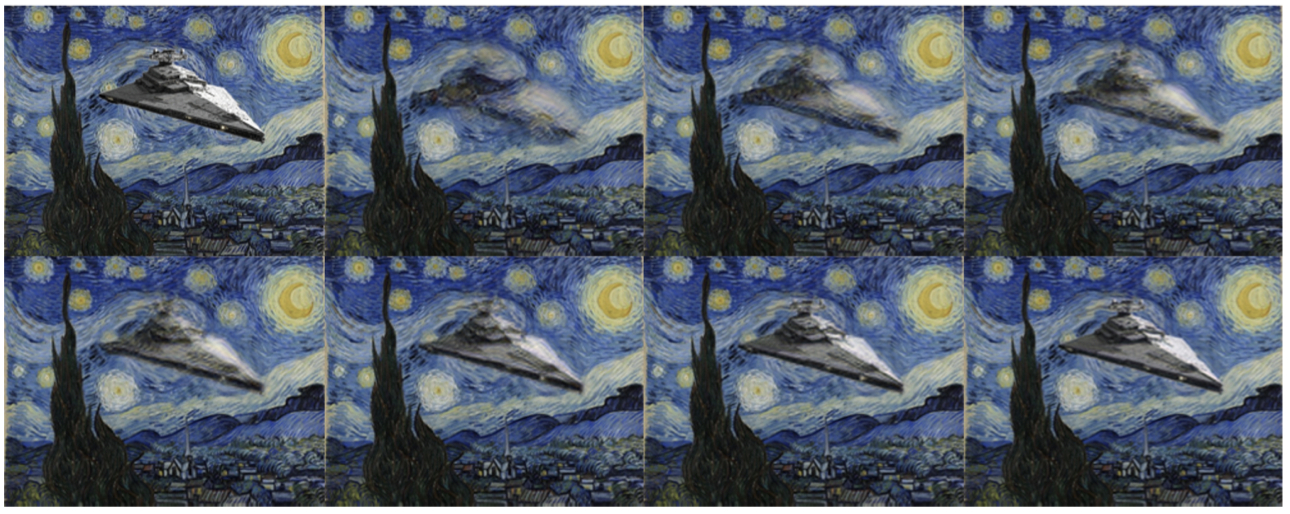}
	\vspace{-4ex}
\caption{Harmonization: Starry-night image with space ship object overlayed. From top-left to bottom-right: (1) Input overlayed image, and (2)-(8) Images generated by injecting image at scales (2) to (8) in the SETGAN model.} 
\vspace{-3.0ex}

    \label{fig:res_harmonization}
\end{figure}

\vspace{0.8ex}

\noindent{\bf Harmonization.} In figure \ref{fig:res_harmonization}, we show the results of using a trained SETGAN model in harmonizing (blending) a spaceship object overlayed on a starry-night image. The images (2)-(8) are generated by injecting images at scales (2) to (8) of the SETGAN model. Based on the scale at which the image is introduced, we find that the degree of blending between the foreground and background image varies. An image generated from coarser scales (2)-(4), both the shape and the texture of the foreground spaceship is matched with the background starry-night painting. However, for the finer scales (5)-(7), the shape of the spaceship is maintained while the texture of the spaceship alone is matched to the background. The energy consumed by the editing process also scales accordingly based on the scale at which the image is injected. Furthermore, in this application, the generative process is done only around the region of the spaceship and the energy consumed is reduced accordingly only for that smaller part of the image.

\begin{figure}[t]
    \centering
	\includegraphics[ width=\linewidth]{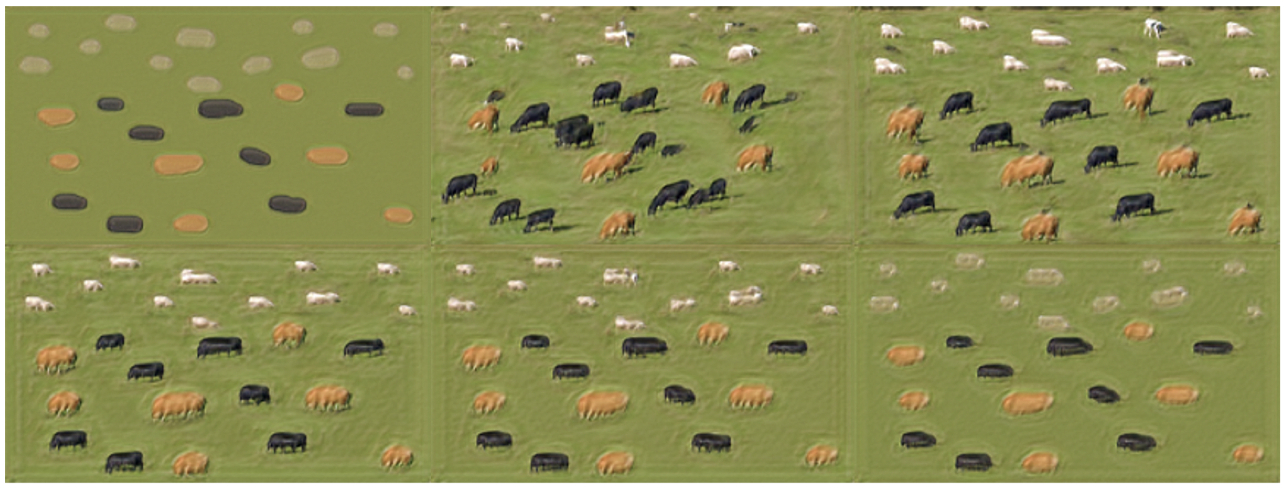}
	\vspace{-4ex}
\caption{Paint2image: Cows image used for clip-art based editing. From top-left to bottom right: (1) Input clip-art image, (2)-(6) Images generated by injecting image at scales (1) to (5) in the SETGAN model.}
\vspace{-3.0ex}

    \label{fig:res_paint2image}
\end{figure}

\vspace{0.8ex}

\noindent{\bf Paint2Image.} In figure \ref{fig:res_paint2image},  we show the results of feeding the clip-art image for the cows' image trained on the SETGAN model. We feed the clip-art image from scales (1) to (5) of the SETGAN model. We can see high-quality features generated by the SETGAN model when injected at coarser scales (1) and (2). 

\begin{figure}[h]
    \centering
	\includegraphics[ width=\linewidth]{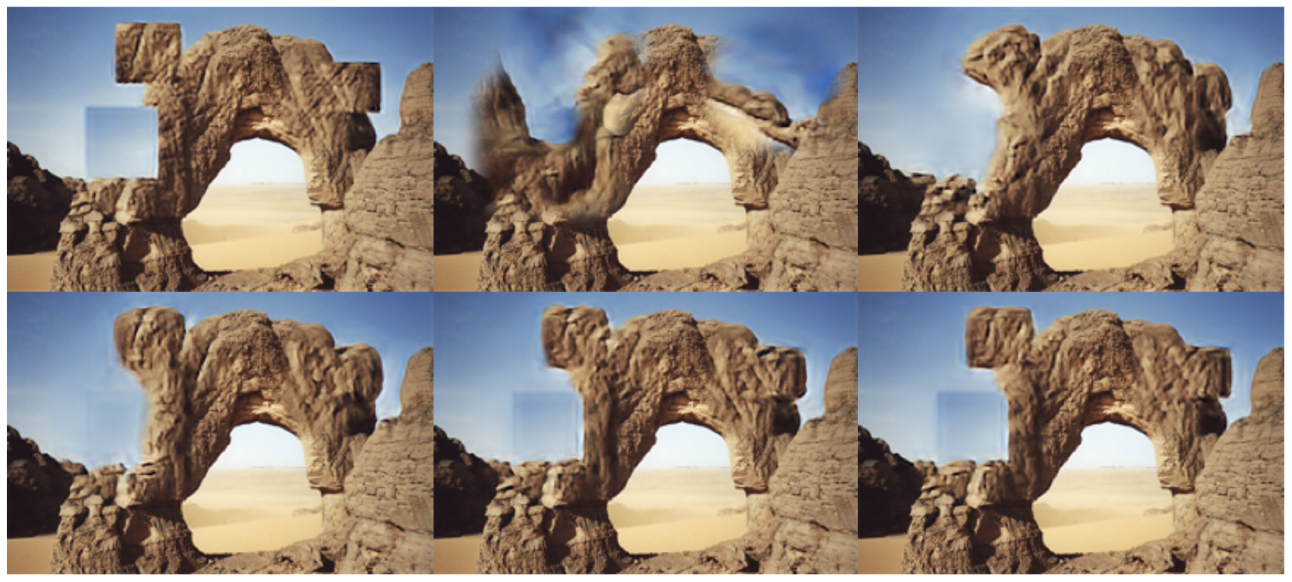}
	\vspace{-4ex}
\caption{Editing: Stones image used for image editing. From top-left to bottom right: (1) Input edited image, (2)-(6) Images generated by injecting image at scales (1) to (5) 
}
\vspace{-2.0ex}

    \label{fig:res_editing}
\end{figure}


\noindent{\bf Editing } In Fig. \ref{fig:res_editing},  we show the results of feeding the edited stone image trained on the SETGAN model. We feed the stone image from scales (1) to (5) of the SETGAN model. For an editing application, feeding the image at the coarsest scale (1) creates artifacts. This is because, at the coarse-scale, it loses the local information. However, higher scales (4)-(5) does not have enough detail. Scales (2)-(3) seems to be the optimal scale for the right set of details. 

From the above results, we can observe that we need to choose the appropriate scale based on the image manipulation task. Fig. \ref{fig:acc_vs_edp} (top) shows the energy consumed by SETGAN across different scales. The energy spent by different applications depends on the scale at which these applications operate. A Paint2Image task operating at the coarsest scale spends the largest energy while the Harmonization task operating at the finer scales consumes the least energy. Since the editing task works at the intermediate scale, it consumes energy between the above two tasks. The energy spent also depends on the size of the image on which the application acts on. Applications like Harmonization and Editing are localized 
and hence, spend less energy when compared to paint2image and super-resolution which work on the whole image. 

\begin{figure}[t]
    \centering
	\fbox{\includegraphics[ width=0.30\linewidth]{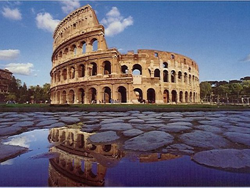}}
	\includegraphics[ width=0.32\linewidth]{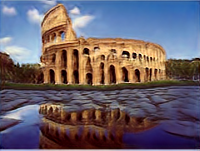}
	\includegraphics[ width=0.32\linewidth]{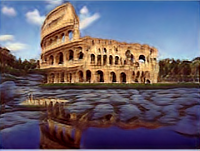}
	\fbox{\includegraphics[ width=0.30\linewidth]{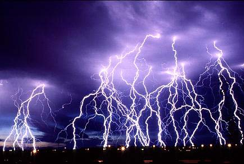}}
	\includegraphics[ width=0.32\linewidth]{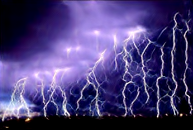}
	\includegraphics[ width=0.32\linewidth]{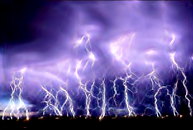}
\vspace{-2.5ex}
\caption{Random images generated using SETGAN model with early exit and parallel training: (Top) Colosseum and (Bottom) Lightning. Leftmost image is the input source image while the next two are the random images.}
\vspace{-2.5ex}

    \label{fig:optimized_rand_img}
\end{figure}

\begin{figure}[t]
    \centering
	\includegraphics[ width=0.24\linewidth]{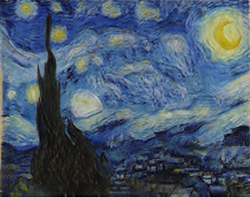}
	\includegraphics[ width=0.24\linewidth]{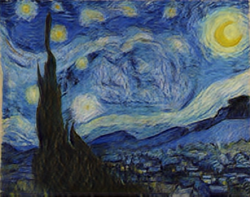}
	\includegraphics[ width=0.24\linewidth]{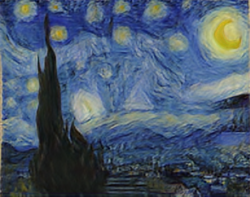}
	\includegraphics[ width=0.24\linewidth]{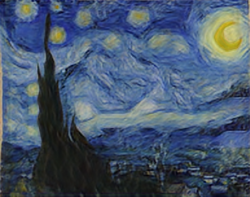}
\vspace{-2.5ex}
\caption{\textcolor{black}{Comparing random images of Starry Night generated using SinGAN (left two) and SETGAN (right two) model}}
\vspace{-2.5ex}

    \label{fig:singan_setgan_comp}
\vspace{-2.5ex}
\end{figure}

\vspace{-1.0ex}

\subsection{Qualitative Analysis}

Fig. \ref{fig:optimized_rand_img} shows some random images generated using SETGAN model. The leftmost image is the source image while the images at center and right are the random images generated using SETGAN model. The overall characteristics of the source image are maintained by our model, but still, the images are different. This can be seen from the number and shape of the windows of the Colosseum image. However, similar to the source image, we see the reflections of the building in the water in all the three images. A similar observation could be made in the number of stars and the swirls in the starry-night image. Fig. \ref{fig:optimized_rand_img} shows the random images generated using the early exit with a reduced number of scales. Even with the reduced number of scales, we see that the images maintain the same characteristics as the un-optimized SETGAN while providing around 56\% reduction in energy. \textcolor{black}{Fig. \ref{fig:singan_setgan_comp} compares images generated using our baseline SinGAN and those generated with SETGAN. We can see that the image characteristics are similar between SinGAN and SETGAN generated images.}

\vspace{-1.5ex}
\section{Conclusions and Future Work}
\vspace{-0.5ex}

We propose and evaluate a new SETGAN framework to trade-off energy and scale of inference using single-image GANs for image manipulation tasks on resource-constrained mobile platforms. The key novelty includes a multi-scale GAN model that can be efficiently trained for each input image using a parallel algorithm and allows efficient inference by automatically selecting the appropriate scale to achieve a user-specified quality threshold. Our results show that optimized SETGAN models can achieve a 56 percent gain in energy for negligible loss in SSIM accuracy and 4x speedup in training when compared to state-of-the-art SinGAN models. Future work includes applying dynamic resource management techniques \cite{TVLSI-2019,TVLSI-2017,TODAES-2020} to further improve the energy-efficiency.
\bibliographystyle{ACM-Reference-Format}
\newpage
\bibliography{files/07_references}

\end{document}